\documentclass[sigconf]{acmart}
\usepackage{booktabs} 
\usepackage{graphicx}
\usepackage[show]{chato-notes}

\usepackage{algorithm}
\usepackage{algorithmic}
\usepackage{tabularx}
\usepackage{calc}

\usepackage{balance}

\usepackage{multirow} 
\usepackage{hyperref}
\usepackage{mathtools}
\DeclarePairedDelimiter\ceil{\lceil}{\rceil}

\setcopyright{rightsretained}

\copyrightyear{2017} 
\acmYear{2017} 
\setcopyright{acmcopyright}
\acmConference{SIGIR '17}{}{August 07-11, 2017, Shinjuku, Tokyo, Japan}
\acmPrice{15.00.}
\acmDOI{http://dx.doi.org/10.1145/3077136.3080811}
\acmISBN{978-1-4503-5022-8/17/08} 

\begin{document}
\title{Deep Character-Level Click-Through Rate Prediction for Sponsored Search}


\author{Bora Edizel}\authornote{This work was done when the author interned at Yahoo.}
\affiliation{%
  \institution{Pompeu Fabra University \& Eurecat}	
  \city{Barcelona, Spain} 
}
\email{bora.edizel@upf.edu}

\author{Amin Mantrach}\authornote{This work was done when the author was at Yahoo.}
\affiliation{%
  \institution{Criteo Labs}
  \city{Palo Alto, California, U.S.A} 
}
\email{a.mantrach@criteo.com}

\author{Xiao Bai}
\affiliation{%
  \institution{Yahoo Labs}
  \city{Sunnyvale, California, U.S.A} 
  }
\email{xbai@yahoo-inc.com}

\fancyhead{}


\begin{abstract}
Predicting the click-through rate of an advertisement is a critical component of online advertising platforms. In sponsored search, the click-through rate estimates the probability that a displayed advertisement is clicked by a user after she submits a query to the search engine. 
Commercial search engines typically rely on machine learning models trained with a large number of features to make such predictions. 
This inevitably requires a lot of engineering efforts to define, compute, and select the appropriate features. 
In this paper, we propose two novel approaches (one working at character level and the other working at word level) that use deep convolutional neural networks to predict the click-through rate of a query-advertisement pair. 
Specifically, the  proposed architectures only consider the textual content appearing in a query-advertisement pair as input, and produce as output a click-through rate prediction. 
By comparing the character-level model with the word-level model, we show that language representation can be learnt from scratch at character level when trained on enough data. 
Through extensive experiments using billions of query-advertisement pairs  of a popular commercial search engine, we demonstrate that both approaches significantly outperform a baseline model built on well-selected text features and a state-of-the-art word2vec-based approach. 
Finally, by combining the predictions of the deep models introduced in this study with the prediction of the model in production of the same commercial search engine, we significantly improve the accuracy and the calibration of the click-through rate prediction of the production system. 
\end{abstract}


\keywords{Deep Learning; NLP; Online Advertising; Sponsored Search; CTR Prediction}

\maketitle

\section{Introduction} 
\label{sec:introduction}
Click-through rate (CTR) prediction is a critical component of any online advertising platform. For an advertisement, the value of the click-through rate can be estimated by the number of times it is clicked divided by the number of times it is shown, quantifying the extent to which an ad\footnote{\small In the remainder, ad(s) will be used to refer to advertisement(s)} is likely to be clicked in a specific context. 
In sponsored search, ad impressions are typically monetized on a pay-per-click basis through the generalized second price auction \cite{NBERw11765}.
Given a query issued by a user, in order to foresee the potential revenue, a commercial search engine has to predict the probability that an ad is clicked by the user for this query (i.e., CTR) as accurately as possible. Over predicting the click-through rate tends to give an ad a higher ranking position in the search result page. If it is not clicked, the search engine does not only lose the expected revenue from this ad but also lose the opportunity of getting more revenue from ads ranked at lower positions due to the position bias affecting ad clicks \cite{chen2012position}. On the other hand, under predicting the click-through rate may result in an ad being placed at a lower position or even not showing up in the search result page, decreasing the revenue that may be made from the ad. 

The problem of click-through rate prediction has led to many research efforts in the past few years \cite{he2014practical, juan2016field}, including those from leading search engine companies \cite{Cheng2010, graepel2010web, mcmahan2013ad}.
So far the most successful models across the industry rely on a large number of well designed features to predict the click-through rate. Despite of the prediction accuracy of such models, it has been noted that it is very challenging to select the right features \cite{he2014practical}, and to deal with feature sparsity and management at scale, etc. \cite{mcmahan2013ad} in a complex dynamic system. 
Moreover, when facing a new context, e.g., new query, new ad, or new query-ad pair, such models may not be able to make accurate predictions \cite{richardson2007predicting}. For instance, the most predictive features of click prediction models are those capturing historical click information \cite{he2014practical} as frequent clicks imply user preference for an ad in a specific context. However, new queries and ads may not have enough history to compute reliable features for accurate prediction. 
To alleviate this kind of cold-start problems, one may rely on hybrid approaches to learn patterns or latent representations from the observed data that generalize well to unobserved contexts, or may rely solely on content-based features that are independent of the click history (see \cite{saveski2014item} for a description of various cold-start solutions). For instance, the BM25 score of an ad relatively to a query  \cite{baeza1999modern} is an effective content-feature of this kind. 
 

Most recently, following the advancements in deep learning, especially in natural language processing, new architectures have been proposed to learn word embeddings and text similarity between a query and a web page \cite{Huang:2013:LDS:2505515.2505665, shen2014latent}, or between a query and an ad \cite{Zhai-2016}. This alleviates the need of designing and implementing large amounts of features, even though maintaining and refreshing the learnt word embeddings still requires huge engineering efforts. 
Interestingly, the last aforementioned advancements in deep learning have not yet been applied to the click prediction problem. Indeed, state-of-the-art CTR prediction models are hybrid models relying on well designed historical, context and content-based features \cite{he2014practical, mcmahan2013ad}.  


 
In this work, we directly predict the click-through rate of a query-ad pair by solely relying on its textual content. 
Specifically, we present two novel deep convolutional neural networks that process the text appearing in a query-ad pair as input without any additional information, and output a CTR prediction. 
The first model learns directly from a binary encoding of the textual input in a bag-of-character space, and does not require any external dictionary. 
The second model exhibits a similar structure but takes as input pre-trained word vectors, and hence assumes a pre-existing word dictionary.   

To the best of our knowledge, we are the first: (1) to learn meaningful textual similarity between two pieces of text (i.e., query and ad) from scratch, i.e., at character level, and (2) to directly predict the click-through rate in the context of sponsored search without any feature engineering. By directly learning and predicting CTR at character level and at word level, we naturally broaden the use of the click prediction model to cold-start (i.e., new) and long-tail (i.e., rare) queries and ads 
as the character-level model can be applied on any query-ad pair as far as their characters are part of the considered input alphabet (e.g., the 26 English letters plus a few punctuations). In fact, although the coverage of the word-level model may be slightly limited by the pre-computed word dictionary, as shown in our experiments, using word-level representation helps to bring external knowledge about the words to boost the prediction accuracy on tail queries and ads. 
 

In this work, we aim at delivering an additional and generalizable signal to improve CTR prediction for sponsored search. We conduct a thorough experimental evaluation, using billions of query-ad pairs from a major commercial search engine, to address the following research questions:

\label{intro:researchquestions}
\begin{enumerate}

\item \emph{Can we automatically learn representations from the query-ad content without any feature engineering in order to predict the CTR in sponsored search?} 
We show that the proposed character-level and word-level deep learning models can improve the AUC of  a feature-engineered logistic regression model with 185 content-based features 
by up to 0.09 (Section \ref{sec:baselines}).
As the three models optimize for the same loss function on the same training data, this clearly shows that the proposed deep models can automatically learn more meaningful representations for predicting CTR of query-ad pairs than 
the feature-engineered model with well-selected features. \\

\item \emph{
How does the performance of the character-level deep learning model differ from that of the word-level model for CTR prediction?} 
In Section \ref{sec:charsvswords}, we show that learning query-ad similarity at character level reaches slightly better performance with an AUC of $0.862$, than its word-level alternative that reaches an AUC of $0.859$. This slight difference is statistically significative on 27M test points. 
Interestingly, character-level model outperforms word-level model when the models are trained with enough data (i.e., more than 1 billion query-ad pairs, Figure \ref{charVsWordFig}). This highlights one of the main findings of this work:  language representation can be learnt from scratch, at character level, without the need of any precomputed dictionary.  In addition, we observe that the word-level model outperforms the character-level model on tails (i.e., queries, ads, and query-ad pairs with low frequency) because it can benefit from the external knowledge provided by the pre-trained word vectors. On the other hand, the character-level model outperforms the word-level model on  heads since it can benefit from the better representations of the domain learnt from scratch. \\


\item \emph{How do the introduced character-level and word-level deep learning models compare to the baseline models? What is the improvement of prediction accuracy on head, torso, and tail queries, ads, and query-ad pairs?}
We show in Section \ref{sec:baselines} that the two proposed models improve the AUC of a baseline model built on well-selected content-based features by up to $0.090$, and the AUC of a word2vec-based approach \cite{grbovic2016scalable} by up to $0.082$ (Table \ref{s2vVsCnnTable}). Specially, the proposed models improve the AUC of the two baselines on head, torso, and tail by up to 0.088,  0.086, and 0.059 respectively (Table \ref{tab:fullheadtorsotail}). \\

%

\item \emph{Can the proposed character-level and word-level deep learning models be leveraged to improve the CTR prediction model running in the production system of a popular commercial search engine?}
In Section \ref{sec:production}, we show that by combining the prediction of one of the deep models proposed in this work (i.e., character-level or word-level model) with the prediction of the production model, we can increase the AUC of the production system by up to 0.86\%. Interestingly, when considering mobile devices, the improvement on AUC reaches 3.95\% (Table \ref{prodVsCnnTable}). 



\end{enumerate}

\section{Related Work}
\label{sec:relatedwork}
In this work, we propose to use deep convolutional neural network to directly learn click-through rate from the characters and the words of query-ad pairs.
We present in this section the state-of-the-art of the various domains covered by this research, and discuss how our contributions differ from the existing works.
We first review the related work in CTR prediction.   
Then, we look at previous research on sentence similarity learning and matching using deep neural networks. 
Finally, we discuss previous deep models working at character level used for tasks different from ours.

\subsection{CTR prediction}
Computational advertisement, and more particularly sponsored search,
has been a subject of study particularly active since the beginning of
the century \cite{mehta2005adwords}.
A large body of work discussing computational advertising is devoted
to finding models and techniques that enable the most accurate prediction
of the probability for an ad to be clicked when returned to a user for her query (i.e., CTR) \cite{graepel2010web, Hillard2011, mcmahan2013ad, richardson2007predicting, Shaparenko2009, Wang2013, ZhangDXFWBWL14}.
Graepel \emph{et al.} \cite{graepel2010web} describe the Bayesian online learning algorithm used in Bing's production system to predict CTR. This model relies on query features, ad features,  context features, as well as the Cartesian product of these base features. 
McMahan \emph{et al.} \cite{mcmahan2013ad} discuss the CTR prediction algorithm used at Google, along with many practical insights to build a large-scale online learning system.  This work particularly confirms the challenges in building CTR prediction model that requires computing, maintaining and serving a large number of sparse contextual and semantic features. 
In fact, even in the related domain of display advertising, machine learning models trained with a large number of features have so far been the mostly adopted. For instance, He \emph{et al.} \cite{he2014practical} present the CTR prediction model at Facebook and clearly point out that the most important challenge to reach accurate predictions is selecting good features, which however may not be trivial. 
There have been different efforts on building features, including text features \cite{Shaparenko2009}, click features \cite{richardson2007predicting}, psychology features \cite{Wang2013}, query segment features \cite{Hillard2011}, to improve CTR prediction models for sponsored search. Most recently, deep neural networks have also been used for CTR prediction. Jiang \emph{et al.} \cite{jiang2016research} proposed to use recurrent neural networks to learn features from queries, ads and clicks for a logistic regression model. Zhang \emph{et al.} \cite{ZhangDXFWBWL14} relies on features extracted from user's sequential ad browsing behavior to train a recurrent neural network. 
Different from all these works, our model does not need any heavy feature engineering but only the textual content appearing in a query-ad pair suffice.  


\subsection{Deep Similarity Learning and Matching}
Matching a search query to a number of ads that are likely to attract user clicks is central to any commercial search engine. With the pervasive success of deep learning, recent works start modeling the similarity between texts using deep network models \cite{arch2_nips} and exploring their use in web search \cite{Huang:2013:LDS:2505515.2505665, shen2014latent} and sponsored search \cite{grbovic2016scalable, Zhai-2016}. 

Grbovic. \emph{et al.} \cite{grbovic2016scalable} mine search sessions that include queries, clicks on ad and search links, dwell times and skipped ads to learn semantic embeddings for queries and ads, and use cosine similarity between the learnt embeddings to measure the similarity between a query and an ad. The main drawback of the approach is that the learning is done at full query level, and ad identifier level. This means that the approach can not exploit two queries with similar content except if they occur often in the same context in search sessions.
The algorithm also suffers from the out-of-vocabulary problem as a significant fraction of search queries are new and advertisers are actively updating their ads.
To solve this problem, Zhai \emph{et al.} \cite{Zhai-2016} propose to use an attention network on top of recurrent neural networks to map both queries and ads to real valued vectors, and then rely on cosine similarity between the query and ad vectors to measure their similarity.  Unlike \cite{grbovic2016scalable}, they are directly working at word level and therefore are less sensitive to the out-of-vocabulary problem.  In this work we actually propose to go down to character level and therefore inherently deal with any input of the considered alphabet. At the difference of  \cite{Zhai-2016}, and  \cite{grbovic2016scalable} we are not learning (in a weakly supervised way) query vectors, and ad vectors to be used in a cosine similarity function but instead are learning (in a supervised way) a complex similarity function embedded in a neural network predicting directly the CTR of a query-ad pair.

In the context of web search, Huang \emph{et al.} \cite{Huang:2013:LDS:2505515.2505665} 
introduce  the letter n-gram based word hashing encoding. 
Compared with the one-hot vector encoding, word hashing allows to represent a query or a document using a vector with much lower dimensionality.
However, when compared to character-level one-hot encodings, the dimension are much higher. Indeed, the character-level encoding dimensionality corresponds to the number of characters of the input times the size of the alphabet, which is much less than the dimensionality of vectors using letter trigrams. Furthermore, this encoding looses the sequence information at the opposite of the character-level one-hot encodings.
Similarly, Shen \emph{et al.} \cite{shen2014latent} use word-n-gram representations of queries and web pages in convolutional neural networks to learn query-document similarity. Different from \cite{Huang:2013:LDS:2505515.2505665}, they project each raw word-n-gram in a low-dimensional feature vector and perform a max pooling operation to select the highest neuron activation value across all word-n-gram features at each dimension. This is similar to the word-level representation used in our deep model. However, our model does not rely on cosine similarity as  \cite{Huang:2013:LDS:2505515.2505665} and \cite{shen2014latent}  but uses the cross-convolutional operator \cite{arch2_nips} to capture query-ad similarity. Another difference between our work and the model proposed in \cite{shen2014latent} is that the latter uses negative sampling on search click logs while we directly use the not-clicked ads as negative samples. 

Hu \emph{et al.} \cite{arch2_nips} also propose to directly capture the similarity between two sentences without explicitly relying on semantic vector representations. 
As DeepWordMatch, this model works at word level, but is targeting matching task as: sentence completion,  matching a response to a tweet, and paraphrase identification.




\subsection{Deep Character-level Models}
There are a number of works learning at character level for different natural language processing (NLP) tasks in recent years. Nogueira dos Santos \emph{et al.} \cite{dos2014learning} are among the first to use character-level information for part-of-speech tagging. They propose to jointly use character-level representation and the more traditional word embedding in a deep neural network for this task. Later on, they propose to use a similar deep neural network with character-level and work-level representations to perform name entity recognition \cite{santos2015boosting}. Unlike these early efforts, our character-level model does not use any word embedding as input. 

Several following works \cite{ballesteros2015improved, DBLP_journals_corr_ConneauSBL16, kim2015character, lecunCharLevelNIPS2015}  demonstrate the power of character-level information alone in NLP tasks. Ballesteros \emph{et al.} \cite{ballesteros2015improved} discuss the benefits of replacing word-level representation by character-level representation in long short-term memory (LSTM) recurrent neural networks to improve transition-based parsing. Kim \emph{et al.} \cite{kim2015character} show in their work that character inputs are sufficient for modeling most of the languages, and their LSTM recurrent neural network language model processing character inputs is as good as the state-of-the-art models using word-level or morpheme-level inputs for English. Zhang \emph{et al.} \cite{lecunCharLevelNIPS2015} explore the use of character-level convolutional networks for text classification and show that character-level convolutional networks  achieve competitive results against traditional models and deep models such as word-based ConvNets \cite{CovNet}. 
Conneau \emph{et al.} \cite{DBLP_journals_corr_ConneauSBL16} further show that when using very deep networks of up to 29 convolutional layers, a model that operates directly at character level achieves significant improvements over the state-of-the-art on several public text classification tasks. Interestingly, in case of big datasets, they report good results using shallower neural networks.

Although character-level models have been successfully applied in so many different tasks, none of them is learning similarity between two pieces of text. This clearly motivates us to design our deep character-level model for click-through rate prediction. 


\section{Deep CTR Modeling} \label{sec:model}
In this section, we design two novel deep convolutional neural networks, namely DeepCharMatch and DeepWordMatch, to directly model the CTR of query-ad pairs based on their content at character level and at word level. 
We start by formalizing the general CTR prediction problem in the context of sponsored search.
We then present the key components of our models. We finally provide details on the input representation and model architecture of the character-level model DeepCharMatch, and the word-level model DeepWordMatch respectively. 

\subsection{CTR Modeling}
\label{sec:modeling}
To model the CTR distribution of query-ad pairs, we have at our disposal a query-ad search log sampled from $\mathcal{QA} \subseteq \mathcal{Q}\times\mathcal{A}$, where $\mathcal{Q}$ is the set of all possible queries that users can submit on their devices (e.g., desktop and mobile), $\mathcal{A}$ is the set of all possible ads that advertisers can register into the advertising platform, and  $\mathcal{QA}$ in the subset of all query-ad pairs that received at least one impression during a time period $\tau$. Each query-ad pair, $q\_a \in  \mathcal{QA}$ is associated to a binary click feedback variable $c_{q\_a} \in \{0,1\}$, 1 meaning clicked and 0 meaning not clicked. 


 \begin{figure}[t]
  \centering
  \resizebox{\columnwidth}{!}{%
    \includegraphics[width=1.0\textwidth]{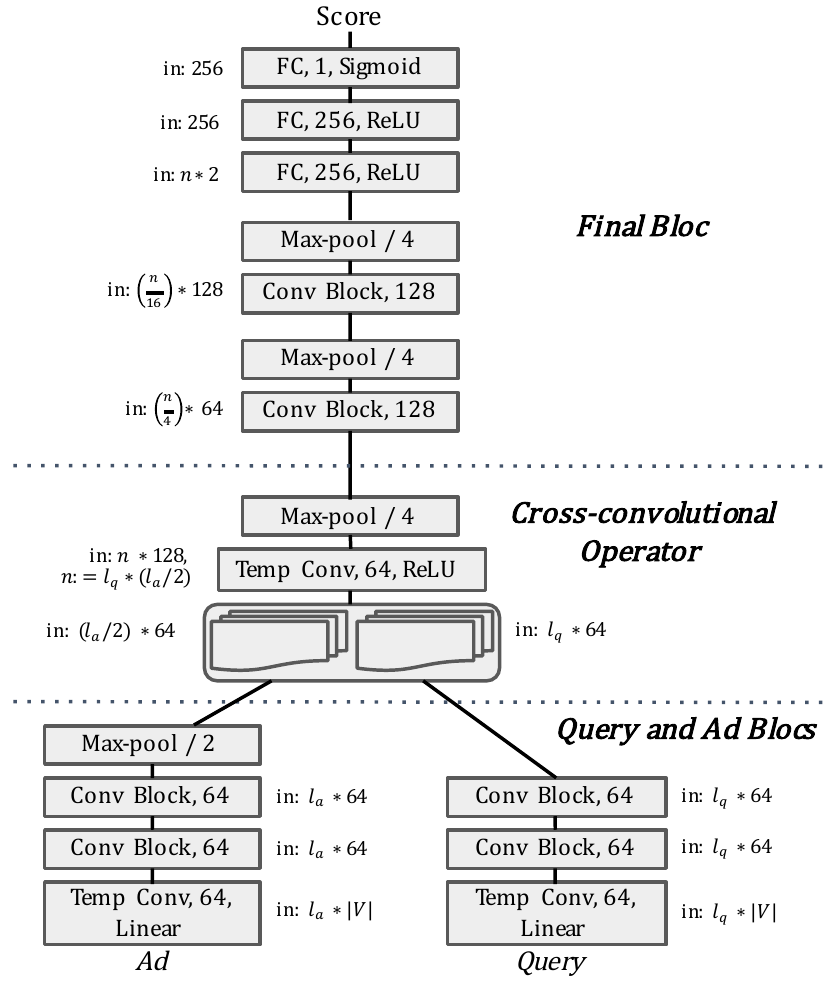}
    }
  \caption{DeepCharMatch Model Architecture.}
  \label{modelArchitecture}
\end{figure}

In order to obtain a well-calibrated 
CTR prediction, we build on the cross-entropy loss function \cite{dreiseitl2002logistic}:

\begin{equation}
\label{eq:logloss}
L = \sum_{q\_a: c_{q\_a}=1} \text{log}\;p_{q\_a} +   \sum_{q\_a: c_{q\_a}=0} \text{log} (1-p_{q\_a}) 
\end{equation} 
where $p_{q\_a} = p(c_{q\_a}=1 | {q\_a}; \theta)$ is the probability for a query-ad pair to be clicked (\textit{i.e.}, the CTR), and  $\theta$ represents the model parameters.
In the remainder, we decompose $\theta$ as parameters of a deep convolutional neural network with the aim of modeling $p_{q\_a}$ directly from the sequence of characters that compose the query and the ad.

\subsection{Key Components}
\textbf{\textit{Temporal Modules.}}
In order to exploit the sequential nature of query and ad at character level, we build on the work of Zhang \emph{et al.} \cite{lecunCharLevelNIPS2015} that introduces the key components to process character-level sequential input in convolutional neural networks.
Since we are dealing with textual data which is one-dimensional and temporal, we make use of temporal convolution, temporal max-pooling and temporal batch normalization. 
These temporal modules work in the same way as their corresponding spatial modules used in images and the only difference is their input dimension.	

\begin{figure}[t]
  \centering
  \resizebox{\columnwidth}{!}{%
    \includegraphics[width=1.0\textwidth]{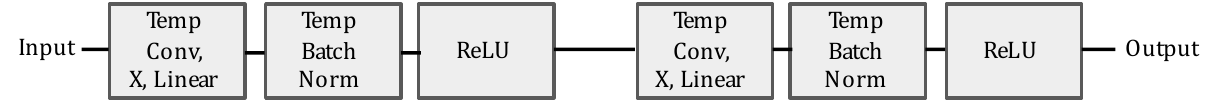}
    }
  \caption{Convolutional Block.}
  \label{convBlock}
\end{figure}

The temporal convolutional module consists of a set of filters whose weights are learnt during the training. 
The module applies a convolution operation between its input and filters. 
Since the filter weights are shared across the input width, patterns can be learnt regardless of locality. 
For the module parameters, we use a fixed filter size of 3, a stride of  1 and we do not use zero paddings. 
In Figures \ref{modelArchitecture} and \ref{convBlock}, we represent convolutions by ``$Temp \text{ } Conv$, $X$, $Y$" where $X$ corresponds to the number of filters and $Y$ corresponds to the activation function.

Temporal max-pooling applies non-linear downsampling to its input in order to reduce dimensionality. 
The downsampling is done by applying a max filter to the non-overlapping partitioned sequences of the initial one-dimensional input. 
Throughout the paper, we refer to temporal max-pooling modules by ``$Max-pool/X$" where $X$ is the size of filter.

Lastly, temporal batch normalization module normalizes its input. 
This accelerates the training obtaining an additional regularization effect \cite{icml2015_ioffe15}.

\textbf{\textit{Convolutional Block.}}
For the ease of the notation, following  \cite{DBLP_journals_corr_ConneauSBL16}, we make use of convolutional blocks (Conv Block). As presented in Figure \ref{convBlock}, a convolutional block is composed of two consecutive sub-blocks where each sub-block is a sequence of a temporal convolution,
a temporal batch normalization, and a ReLU activation function \cite{glorot2011deep}. ReLU is a non-saturating activation function such that for an input $x$, it outputs $max(0,x)$.

\textbf{\textit{Other functions.}}
Additionally, we use fully connected layers. We refer to them by ``$FC$, $X$, $Y$" where $X$ is the number of neurons and $Y$ is the activation function. 

\subsection{DeepCharMatch Model}
The architecture of the character-level CTR prediction model is presented in Figure \ref{modelArchitecture}.
We refer to this model by DeepCharMatch. Before exploring its architecture, we first explain how the input is represented at character level. 

\subsubsection{Input Representation}
Considering an alphabet $V$, and fixed query length $l_q$, queries are represented with a matrix of one-hot-encodings (\textit{i.e.}, binary) of size $l_q \times |V|$. 
More precisely, each character of the query sequence corresponds to a row of size $1 \times  |V|$ in the input matrix. 
Each row contains only one unique entry that is set to $1$ at the position corresponding to the dimension indicated by the considered character of the query while all the other entries of the same row are set to $0$. Hence, the full query matrix has a number of $1$ that corresponds to the length of the query, \textit{i.e.}, $l_q$ (and thus to number of rows of the matrix). 
By so doing, the $i^{th}$ row of the matrix encodes the $i^{th}$ character of the query.
When the query length is smaller than $l_q$ we use zero-paddings, \textit{i.e.}, the remaining rows are fully padded with zeros entries. 
When the query length is larger than $l_q$, we simply ignore all the characters appearing after the $l_q^{th}$ character of the query. 

The same approach is used for representing ads. The three components of a textual ad, \textit{i.e.}, ad title,  ad description and ad display URL, are first concatenated in the aforementioned order in one unique sequence, and then encoded with a matrix of of one-hot-encodings of size $l_a \times |V|$.

\subsubsection{Model Architecture}
DeepCharMatch consists of two parallel deep architectures that are joined by a cross-convolutional operator followed by a final bloc that models the relationship between a query and an ad. We detail this architecture in the following. 

\textbf{\textit{Query and Ad Blocs.}}
Query and ad blocs are two parallel structures that take as input a character-level one-hot encodings of the query and the ad respectively.
Each bloc is a sequence of a temporal convolution, followed by 2 convolutional blocks whose output is a vector representation of the considered input (\textit{i.e.}, the query or the ad).
Learnt representations can be seen as a higher-level representation of the query and of the ad. 

\textbf{\textit{Cross-convolutional Operator.}}
Cross-convolutional operator (as introduced in \cite{arch2_nips}) takes as input higher-level representations of the query and of the ad (outputs of query and ad blocs) and operates a convolution on the cross-product of the query and the ad. 
More precisely, let $H_Q$ be the higher-level query matrix representation with dimensions $k \times l$ and 
$H_A$ be the higher-level ad matrix representation with dimensions $m \times r$ and 
$H_{AQ}$ be the cross product of $H_Q$ and $H_A$ with dimensions $(k*m) \times (l+r)$. Formally, each row of $H_{AQ}$ is set to
$$ H_{QA}[i, \cdot] :=  H_Q[ \ceil*{i / m}, \cdot ]^\frown  H_A[i-m*(\ceil*{i/m}-1), \cdot],$$

\noindent where $i \in \mathbb{N}$, $1\leq i \leq k*m$ and $^\frown$ represents concatenation.
With this operation, we aim to capture possible intra-word and intra-sentence relationships between the query and the ad. 
A temporal max-pooling is applied at the end of this operation.

\textbf{\textit{Final Bloc.}}
The final operations start with a sequence of two blocks, where each block is a convolutional block followed by a temporal max-pooling. Finally, the architecture is ended with three fully connected layers. The Final Bloc models the relationship between the ad and the query.
The output of the final bloc is $p_{q\_a}$ 
which is the CTR prediction of DeepCharMatch for the query-ad pair $q\_a$.

\subsection{DeepWordMatch Model}
\label{par:deepwordmatch}
We also propose a deep convolutional neural network using word-level input. We refer to this model as DeepWordMatch. DeepWordMatch is also trained to maximize the conditional log-likelihood of the clicked and non-clicked sponsored impressions, \textit{i.e.}, ads (Equation \ref{eq:logloss}), and hence outputs the CTR prediction of the considered query-ad pair $q\_a$.

\subsubsection{Input Representation}
Different from DeepCharMatch, DeepWordMatch processes pre-trained, word vectors instead of one-hot character encodings as input. The word vectors can be learnt using either search logs or external sources like Wikipedia \cite{pennington2014glove}. Devising the best way of training word vectors is an interesting open problem but is independent of the model we propose. Considering given word dictionary $W$,  dimension of word vectors $d_w$, fixed query length  $d_q$ and fixed ad length $d_a$,  
queries are represented with a query matrix with dimension $d_q \times d_w$. Similarly,  ads are represented with an ad matrix with dimension $d_a \times d_w$. 

\subsubsection{Model Architecture}
The structure of the neural network is inspired from the matching algorithm for natural languages sentences introduced in \cite{arch2_nips}.
It consists of a cross-convolution operator ended by a final block capturing the commonalities between the query and the ad. Ad and query matrixes consist of pre-trained word vectors directly feed into cross-convolution operator. In order to control dimensionality, kernel sizes of the temporal max-pooling operations are set to 2. Except those points, the architecture of DeepWordMatch is equivalent to the architecture of DeepCharMatch.

\section{Experiments} \label{sec:exp}
\label{sec:experiments}
\begin{figure*}[h]
  \centering
    \includegraphics[width=1\textwidth]{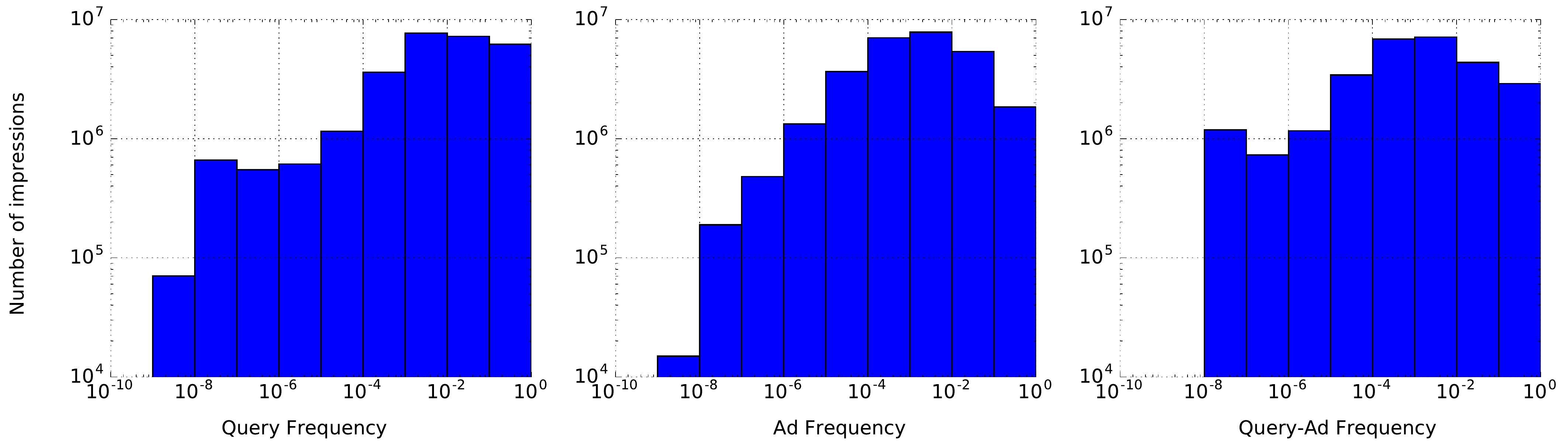}
  \caption{Distribution of impressions in the test set with respect to query, ad, and query-ad frequencies computed on six months (The frequencies are normalized by the maximum value in each subplot).}
  \label{distributionsFig}
\end{figure*}
We evaluate the performance of the proposed CTR prediction models in this section.
We first present how we collect the data set to conduct the experiments, the different baselines that are important to this study, and the metrics that are relevant to evaluate CTR prediction models. We also describe the platform to run our experiments and the choice of parameters.
We then dive into each research question raised in the introduction and discuss the results we obtain from the related experiments.

\subsection{Experimental Setup}

\subsubsection{Dataset}
In order to test our research hypotheses (Section \ref{intro:researchquestions}) we randomly sample query-ad pairs served by a popular commercial search engine. More precisely, we randomly sample from the log the \emph{training set} that consists of about 1.5 billion query-ad pairs served during the period going from August 6 to September 5, 2016.  We only consider the sponsored ads that are shown in the north of the search result pages (\textit{i.e.} above the algorithmic search results). Each sampled query-ad consists of the query,  the ad title, the ad description and the displayed URL of the ad's landing page, in their canonical form, and a binary variable indicating if the ad is clicked or not. 
In the following 15 days from September 6 to September 20, 2016, We randomly sample the \emph{test set} that consists of about 27 millions query-ad pairs without any page position restriction (\textit{i.e.}, we also test for the ads displayed on the east and the south of search result pages).

In order to study the performance of our models on queries and ads with different popularity, we compute query frequency distribution, ad frequency distribution and query-ad frequency distribution of the queries and ads in our test set over a long period (\textit{i.e.}, some consecutive months of 2016). Figure \ref{distributionsFig} reports the distributions of the total number of ad impressions related to the queries, ads or query-ad pairs following into each frequency bin. 
Notice that this period fully covers the periods in which our training and test sets are generated. Therefore, low frequency measures the coldness relative to the entire period. 

\begin{table}[t]
\caption{AUC of DeepCharMatch, DeepWordMatch, Search2Vec and FELR.}
\label{s2vVsCnnTable}
\centering
\begin{tabular}{ll|ll}
                   & All   & Desktop  & Mobile      \\ \hline
DeepCharMatch & \textbf{0.862}  & \textbf{0.870}          & \textbf{0.828}              \\
DeepWordMatch & 0.859  & 0.867          & 0.827                \\
Search2Vec         & 0.780 & 0.796 & 0.705 \\
FELR & 0.772 & 0.784 & 0.710
\end{tabular}
\end{table}

\begin{figure*}[t]
  \centering
    \includegraphics[width=1\textwidth]{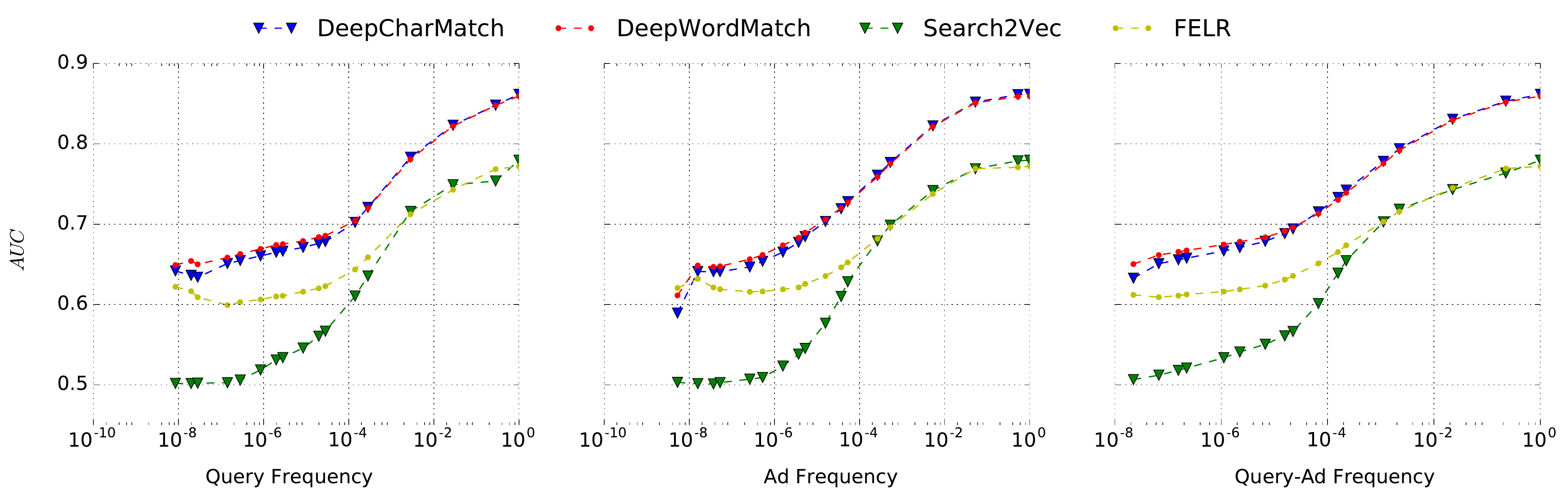}
  \caption{Cumulative AUC by query, ad, and query-ad frequency for DeepCharMatch, DeepWordMatch, Search2Vec and FELR. Frequencies are normalized by the maximum value in each subplot. For each bin, the number of  impressions used to compute AUC  is reported in Figure \ref{distributionsFig}. Cumulative means that at $x$ the plot reports AUC of points whose frequency is lower than $x$.}
  \label{s2vVsCnnFig}
\end{figure*}
\begin{table*}[t]
\centering
\caption{AUC of DeepCharMatch, DeepWordMatch, Search2Vec and FELR, on \emph{tail}, \emph{torso}, and \emph{head} of the query, ad, and query-ad frequency distributions. Tail stands for normalized frequency $nf<10^{-6}$, torso for $10^{-6}\le nf < 10^{-2}$, and head for $nf\ge 10^{-2}$.}
\label{tab:fullheadtorsotail}
\begin{tabular}{cccc|ccc|ccc}
              & \multicolumn{3}{c|}{Query} & \multicolumn{3}{c|}{Ad} & \multicolumn{3}{c}{Query-Ad} \\
              & \emph{tail}    & \emph{torso}    &  \emph{head}   & \emph{tail}   & \emph{torso}   &  \emph{head}  & \emph{tail}     & \emph{torso}    &  \emph{head}    \\ \cline{2-10} 
DeepCharMatch & 0.661   & \textbf{0.814}   & \textbf{0.909}  & 0.659  & \textbf{0.836}  & \textbf{0.926} & 0.665    & \textbf{0.828}   & \textbf{0.943}   \\
DeepWordMatch & \textbf{0.670}   & 0.812   & 0.907  & \textbf{0.668}  & 0.835  & 0.922 & \textbf{0.674}    & 0.826   & \textbf{0.943}   \\
Search2Vec    & 0.521   & 0.739   & 0.817  & 0.516  & 0.753  & 0.844 & 0.532    & 0.740   & 0.854   \\
FELR          & 0.606   & 0.733   & 0.821  & 0.618  & 0.751  & 0.830 & 0.615    & 0.742   & 0.879  
\end{tabular}
\end{table*}
\subsubsection{Baselines}\label{sec:baselines}

\sloppy \textbf{\textit{Feature-engineered logistic regression (FELR).}}  Logistic regression is a state-of-the-art algorithm to predict CTR at massive scale \cite{mcmahan2013ad}.
Therefore, we implement a logistic regression model with content-based features as a baseline. 
Our objective is to test the hypotheses that DeepCharMatch and DeepWordMatch can learn directly from the textual input meaningful representations that are better than feature-engineered models.
The logistic regression model optimizes the same cross-entropy loss function as DeepCharMatch and DeepWordMatch (Equation \ref{eq:logloss}). 
In this case, $\theta$ is simply a parameter vector representing the weights to be learned for each feature along with the bias. 
We use the 185 state-of-the-art features designed to capture the pairwise relationship between a query and the three different components in a textual ad, \textit{i.e.},  its title, description, and display URL. The full set of features consists of 12 common counts features, 12 Jaccard features, 10 length features, 4 cosine similarity features, 4 BM25 features, 8 Brand features, 4 LSI features, 3 semantic coherence features, and 128 hash embedding features. These features are explained in details in \cite{Aiello-2016} and are achieving state-of-the-art results in relevance prediction for sponsored search. Most of these features are the state-of-the-art features in traditional search tasks as supervised ranking, semi-supervised ranking, and ranking aggregation \cite{QinL13}.

\textbf{\textit{Search2Vec.}} Our second baseline is a state-of-the-art word2vec-based approach, namely Search2Vec \cite{grbovic2016scalable}, which learns semantic embeddings for queries and ads from search sessions, and uses the cosine similarity between the learnt vectors to measure the textual similarity between a query and an ad. This approach leads to high-quality query-ad matching in sponsored search. Different from DeepCharMatch and DeepWordMatch, Search2Vec does not learn CTR directly. Instead, clicks are used indirectly in the session data as context of the surrounding queries and ads. Therefore, this approach is considered to be weakly-supervised. Another important difference is that Search2Vec works at query level and ad level, implying that it is more sensitive to the out-of-vocabulary problem.

\textbf{\textit{Production model.}} As a very strong baseline, we are considering the CTR prediction model in the production system of a popular commercial search engine. This model is a machine learning model trained with a rich set of features, including click features, query features, ad features, query-ad pair features, vertical features, contextual features such as geolocation or time of the day, and user features. The learning algorithm is optimizing Equation \ref{eq:logloss} as well.  This model involves great engineering efforts to design relevant features, especially those content-based features extracting the relationship between queries and ads. 
Our objective is to study what are the relative improvements one can expect in production 
when adding a deep learning content-based dimension (\textit{i.e.}, DeepCharMatch or DeepWordMatch prediction) into this model.
Therefore, we use a simple approach that averages the deep model CTR with the production model CTR.
In the remainder we refer to these  algorithms  as DCP and DWP, for the combination of character-level model, and word-level model respectively with production.
While these approaches are very simple, it suffices to demonstrate that a content-based deep learning approach can be leveraged to improve the model in production of a commercial search engine that is not learning automatically content-based query-ads representations.

\subsubsection{Evaluation Metrics}
We measure two standards metrics:  (1) the area under the ROC curve (AUC), and (2) the calibration of the CTR \cite{baeza1999modern}.

\textbf{\textit{AUC.}} For comparing the different baselines we use the AUC to evaluate the ability of the different methods to predict which ad impressions are going to be clicked.
On ad impressions held out during learning, AUC measures whether the clicked ad impressions are ranked higher than the non-clicked ones.
The perfect ranking has an AUC of 1.0, while the average AUC for random rankings is $0.5$. 

\textbf{\textit{Calibration.}} The calibration is the ratio of the number of expected clicks to the number of actually observed clicks. Having a well calibrated prediction of CTR insures that advertisers are paying a fair price, and is thus critical for online adverting auction. The closer the calibration measure is to 1.0, the better the CTR prediction is \cite{he2014practical}.


\subsubsection{Experimental Platform}
We train DeepCharMatch, DeepWordMatch, and FELR using the same environment.
We use the distributed Tensorflow platform\footnote{\small https://www.tensorflow.org/} in an asynchronous fashion on multiple GPUs. 
Adam Optimizer~\cite{kingma2014adam} is used in order to optimize the cross-entropy loss function. 
To initialize the parameters, we use the initialization strategy described in \cite{he2015delving}. 
The mini-batch size is set to 64.
For DeepCharMatch, we fix the query  length $l_q$ to 35 characters, and ad length $l_a$ to 140 characters. 
For DeepWordMatch, we fix the query word-length $d_q$ to 7 and the ad word-length $d_a$ to 40. 
The word vectors feeding the input of DeepWordMatch are publicly available\footnote{\small http://nlp.stanford.edu/data/glove.6B.zip} and consists of 50 dimension vectors obtained by running GloVe algorithm on Wikipedia and Gigaword5\footnote{\small https://catalog.ldc.upenn.edu/LDC2011T07}
\cite{pennington2014glove}.
\subsection{Experimental Results}
We report in this section the performance of the proposed character-level and word-level CTR prediction models by answering the following research questions. 

\paragraph{Research Question 1} 
\emph{Can we automatically learn representations from the query-ad content without any feature engineering in order to predict the CTR in sponsored search?} 

Our objective is to test the extent to which the representations learnt by the introduced deep models are more effective than the 185 engineered-features injected in a state-of-the-art algorithm for CTR prediction, \textit{i.e.}, FELR. 
We show that DeepCharMatch and DeepWordMatch outperform FELR on Desktop and Mobile in terms of AUC by up to 0.086  and 0.118 respectively  (Table \ref{s2vVsCnnTable}). 
This confirms that both models can automatically learn  more effective query-ad representations for predicting the CTR than the 185 engineered-features used by FELR.
Notice that the improvements are larger on the head of the distributions where the frequency are the highest (Figure \ref{s2vVsCnnFig} and Table \ref{tab:fullheadtorsotail}). This highlights that the automatically learnt representations 
generalize better than
engineered features when queries, ads, and query-ads are more frequent.
\label{sec:baselines}




\label{sec:charsvswords}
\paragraph{Research Question 2}
 \emph{How does the performance of the character-level deep learning model differ from that of the word-level model for CTR prediction?} 

Here, we are interested to study the learning curve performance of the two proposed models. 
In other words, as we increase the number of training samples, what AUC can we expect on the same independent randomly sampled test set. 
We hypothesize that by learning at character level instead of word level, the model can fit the CTR distribution better. This hypothesis is based on the fact that the character-level model has a higher degree of freedom as it needs to learn 1,672,002 parameters, \textit{i.e.}, 16.21\% more than the word-level model (Table \ref{numOfParam}). 

\begin{table}[!h]
\centering
\caption{Number of parameters per model}
\label{numOfParam}
\begin{tabular}{cccc}
DeepCharMatch & DeepWordMatch & FELR \\ \hline
 1,672,002       & 1,438,786       & 186               
\end{tabular}
\end{table}

To answer this question, we conduct a parallel experiment between DeepCharMatch and DeepWordMatch. 
Both models are studied under exactly the same conditions, \textit{i.e.},  learnt on the same training points and assessed on the same test points.
 
\begin{figure}[t]
  \centering
  \resizebox{\columnwidth}{!}{%
    \includegraphics[width=1\textwidth]{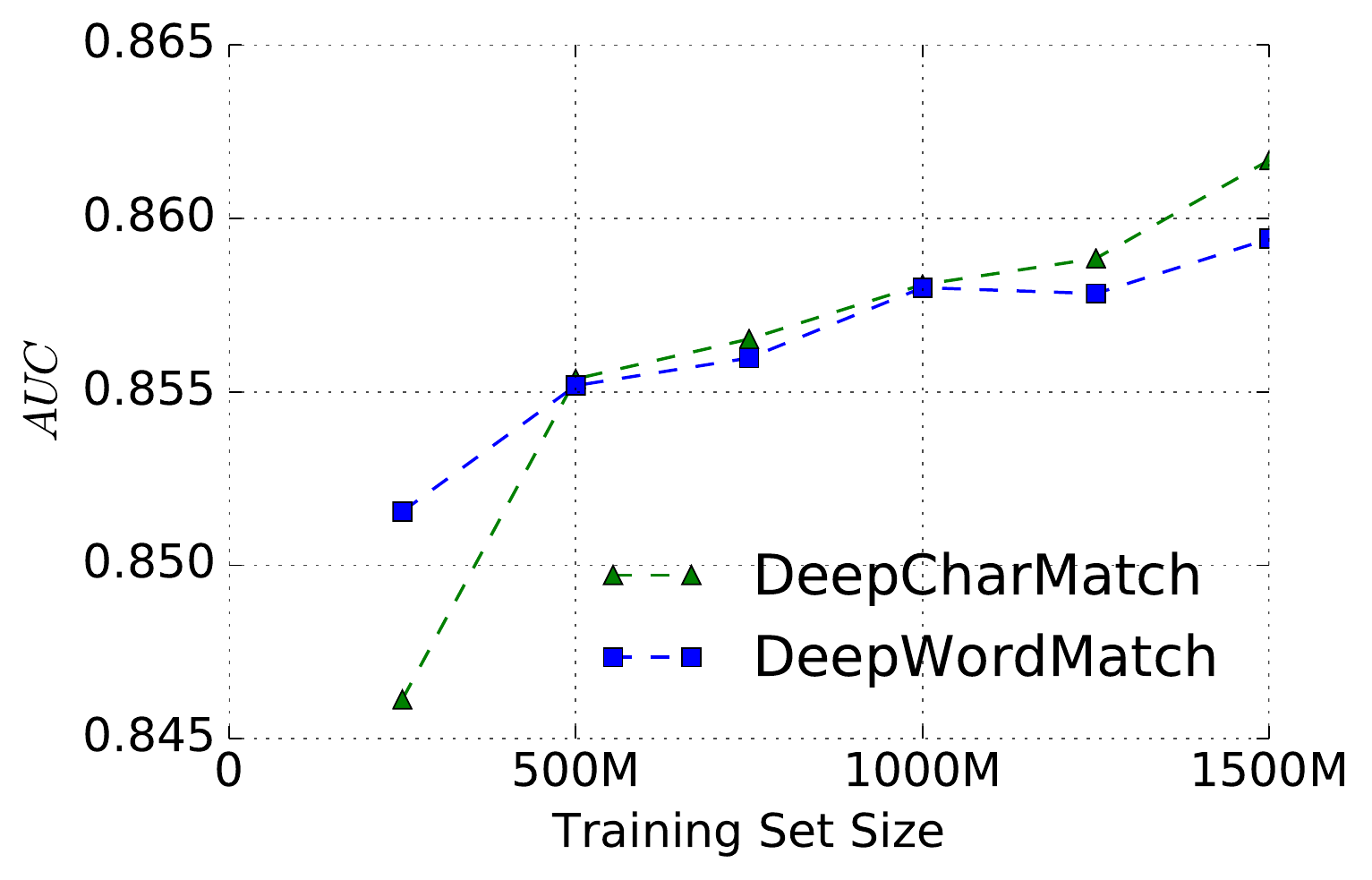}}
  \caption{AUC of DeepCharMatch and DeepWordMatch by number of training points.} 
  \label{charVsWordFig}
\end{figure}

We show that as the number of training points goes over one billion, DeepCharMatch offers more flexibility than DeepWordMatch to learn the underlying query-ad matching. At 1.5 billion points the difference becomes significant: on the 27 millions test points, DeepCharMatch and DeepWordMatch reaches an AUC of 0.862 and 0.859 respectively  (Figure \ref{charVsWordFig} and Table \ref{s2vVsCnnTable}). 
This confirms that, when provided with enough training points, DeepCharMatch learns the underlying representations from scratch without the need of a pre-computed dictionary (like the Wikipedia word vectors used by DeepWordMatch in this particular study).
Interestingly, when looking at the differences in terms of heads and tails of the query, ad, and query-ad distributions, DeepCharMatch outperforms or equals to DeepWordMatch on the heads where the volume is the highest. Inversely, DeepWordMatch outperforms DeepCharMatch on the tails (Table \ref{tab:fullheadtorsotail}). Indeed, DeepWordMatch has an advantage on the tails as it benefits from pre-trained vectors that provide at cold-start an initial knowledge that the character-level model is ignorant of. On the other hand, when shifting towards the heads, DeepCharMatch benefits from its representations learnt from scratch to build a better understanding of the domain.

\paragraph{Research Question 3}
 \emph{How do the introduced character-level and word-level deep learning models compare to the baseline models? What is the improvement of prediction accuracy on head, torso, and tail queries, ads, and query-ad pairs?}
 
Here we are interested to study how the proposed deep models compare to both baselines FELR and Search2Vec especially in different areas of the query, ad, and query-ad distributions. 
While globally Search2Vec outperforms FELR (Table \ref{s2vVsCnnTable}), it appears to be a very poor baseline when it comes to tail queries, ads, and query-ads (Table \ref{tab:fullheadtorsotail}). This emphasizes the fact that Search2Vec is working at query, and ad-identifier level. In other words, the approach cannot relate similar ads or queries based on their content, but only based on their occurrences in the same context. Both DeepCharMatch and DeepWordMatch consistently outperform the baselines from a minimum 0.041 of AUC on the tail of the ad distribution, to a maximum of 0.088 of AUC on the head of the queries distribution.
Interestingly, the improvements over the baselines are more important for mobile devices (Table \ref{s2vVsCnnTable}), which could indicate that the introduced deep models are less sensitive to the sampling bias (as our training data is by nature more populated with desktop impressions). We leave as further work a deeper analysis of the importance of the sampling on performance obtained by each device.



\begin{figure*}[t!]
  \centering
    \includegraphics[width=1\textwidth]{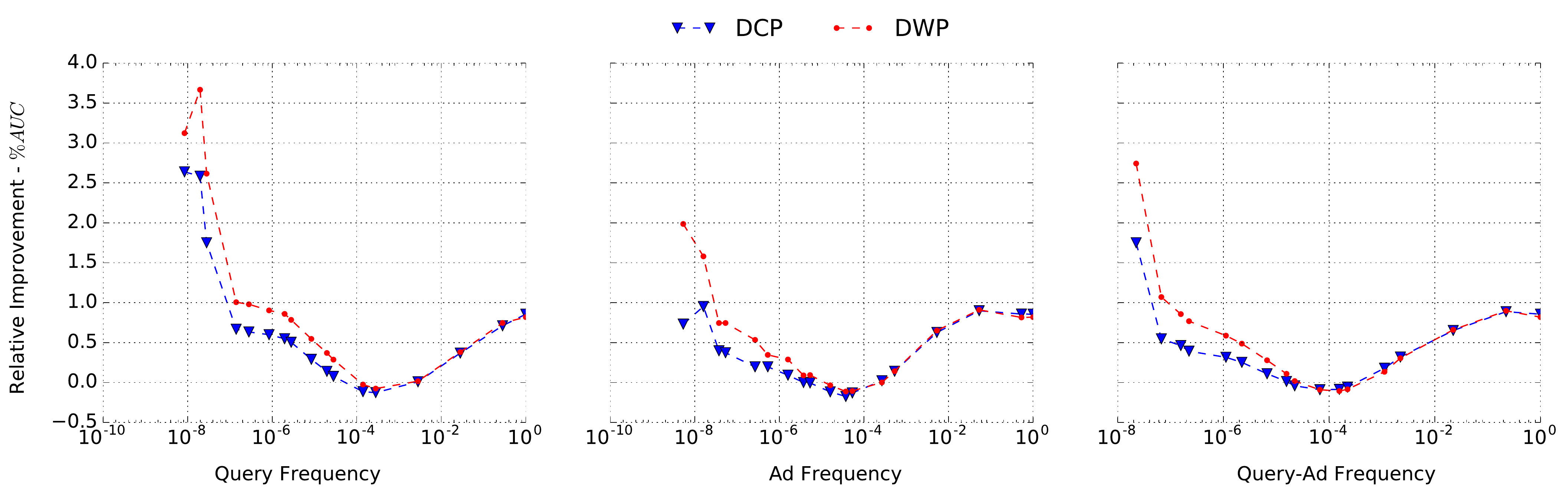}
  \caption{Cumulative relative improvements of DCP and DWP over Production model in terms of \%$AUC$.  Frequencies are normalized by the maximum value of each subplot. For each bin, the number of  impressions used to compute AUC  is reported in Figure \ref{distributionsFig}. Cumulative means that at $x$ the plot reports relative improvements of points whose frequency is lower than $x$.}
  \label{prodVsCnnFig}
\end{figure*}

\paragraph{Research Question 4}
\label{sec:production}
\emph{Can the proposed character-level and word-level deep learning models be leveraged to improve the CTR prediction model running in the production system of a popular commercial search engine?}

The current system in production is relying on engineering-efforts to design relevant content features extracting the relationship between queries and ads.
Our objective is to study how much relative improvements one can expect in production in terms of AUC and calibration when adding a deep learning content-based dimension into the production model. 
We proceed this test using two simple approaches, DCP and DWP,  which average respectively the character-level and word-level predicted CTR to the production predicted CTR (as explained in Section \ref{sec:baselines}).
We observe relative improvements of 0.86\% of  the production AUC when considering the complete test set, with very significant improvements of 3.75\% of the production AUC when restricting to mobile devices (Table \ref{prodVsCnnTable}). Major gains are observed on the head of the distribution with improvements of up to 1.18\% of the production AUC (Table \ref{tab:aucimprovdpcprod}). Interestingly, as observed previously, we observe higher improvements with the character-level model at the heads of the distributions, while the word-level model is more beneficial on the tails. This highlights that DeepWordMatch benefits more of its pre-trained vectors on tails, while on the heads the DeepCharMatch reaches a better understanding of the domain by learning directly its representations at character level. 

\begin{table}[t]
\centering
\caption{Relative AUC Improvement in \% of DCP over Production model.}
\label{prodVsCnnTable}
\begin{tabular}{c|ccc}
    & All & Desktop & Mobile \\ \hline
DCP & \textbf{0.86}  & \textbf{0.29}   & 3.76  \\
DWP & 0.82  & 0.23   & \textbf{3.95} 
\end{tabular}
\vspace{-0.2cm}
\end{table}

\begin{table}[t]
\centering
\caption{Relative Calibration Improvement in \% of DCP over Production Model. }
\label{calibRelImp}
\begin{tabular}{c|ccc}
    & All & Desktop & Mobile \\ \hline
DCP & \textbf{35.76} & \textbf{34.95}  & \textbf{40.40} \\
DWP & 32.21 & 30.85  & 38.50
\end{tabular}
\end{table}

\begin{table*}[h!]
\centering
\caption{Relative AUC Improvements  in \% of DCP and DWP over Production on \emph{tail}, \emph{torso}, and \emph{head} of the query, ad, and query-ad frequency distributions. Tail stands for normalized frequency $nf < 10^{-6}$, torso for $10^{-6}\le nf < 10^{-2}$, and head for $nf\ge 10^{-2}$.}
\label{tab:aucimprovdpcprod}
\begin{tabular}{cccc|ccc|ccc}
    & \multicolumn{3}{c|}{Query} & \multicolumn{3}{c|}{Ad} & \multicolumn{3}{c}{Query-Ad} \\
    & tail    & torso  & head   & tail   & torso  & head  & tail     & torso    & head    \\ \cline{2-10} 
DCP & 0.593   & 0.205  & \textbf{1.176}  & 0.127  & 0.793  & \textbf{0.817} & 0.322    & \textbf{0.584}    & 1.010   \\
DWP & \textbf{0.906}   & \textbf{0.218}  & 1.096  & \textbf{0.324}  & \textbf{0.818}  & 0.723 & \textbf{0.604}    & 0.571    & \textbf{1.090}  
\end{tabular}
\end{table*}


\begin{table*}[h]
\centering
\caption{Relative Calibration Improvements in \% of DCP  and DWP over Production on \emph{tail}, \emph{torso}, and \emph{head} of the query, ad, and query-ad frequency distributions.}
\label{tab:calibimprovdpcprod}
\begin{tabular}{cccc|ccc|ccc}
    & \multicolumn{3}{c|}{Query} & \multicolumn{3}{c|}{Ad}  & \multicolumn{3}{c}{Query-Ad} \\
    & tail    & torso   & head   & tail   & torso  & head   & tail     & torso   & head    \\ \cline{2-10} 
DCP & \textbf{31.44}  & \textbf{33.62}  & \textbf{38.02} & \textbf{32.82} & \textbf{35.01} & \textbf{38.38} & \textbf{30.28}   & \textbf{34.30}  & \textbf{42.64}  \\
DWP & 30.73  & 31.02  & 32.96 & 29.89 & 32.27 & 32.31 & 29.33   & 31.51  & 37.68 
\end{tabular}
\end{table*}

For calibration analysis, let us consider the relative gain in calibration of model $\mathcal{M}$ over the Production model:
 \begin{equation}\nonumber
 \frac{|1-Calibration(Production)| - |1-Calibration(\mathcal{M})|}{|1-Calibration(Production)|}
 \end{equation}
This gain measures the relative decrease of the calibration error in production (\textit{i.e.} |1-Calibration(Production)|) when using model  $\mathcal{M}$.  
We observe an important relative calibration gain in production  with DCP model and DWP model with up to 34.95\% on Desktop and 40.40\% on  Mobile (Table \ref{calibRelImp}). Finally, we observe that the calibration is better on the heads than on the tails especially with the character-level model (Table \ref{tab:calibimprovdpcprod}).

\section{Conclusions} \label{sec:conclusion}
\label{sec:conclusions}
We present in this paper two new content-based click-though-rate prediction models for sponsored search. 
Both models are built on convolutional neural network architectures and learnt in a supervised way from clicked and non-clicked query-ad impressions sampled from the log of popular a commercial search engine.
We demonstrate through large-scale experiments (with 1.5 billions query-ad training samples) that query-ad representations can be learnt from scratch, at character level, to predict the CTR, and the prediction is particularly accurate for frequent queries, ads and query-ad pairs. We also show that when using pre-trained word vectors, the proposed word-level model can make more accurate prediction on the tail of the query, ad and query-ad frequency distributions than the character-level model.
One important contribution of this work is to show that predicting CTR of query-ad pairs directly at character level can outperform the traditional machine learning models trained with well-designed features. Particularly, when 
combining the CTR prediction of the proposed deep learning models with that of the machine learning model trained with a rich set of content-based and click-based features in the production system of a popular commercial search engine,
we can significantly improve the accuracy and the calibration of the model in production.

\footnotesize
\balance
\bibliographystyle{ACM-Reference-Format}
\bibliography{sigproc}

\end{document}